\documentstyle[aaai,named]{article}
\nocopyright
\newtheorem{theorem}{Theorem}

\newtheorem{define}{Definition}
\newtheorem{example}{Example}
\newcommand{\nl}{\par\noindent}
\newcommand{\ifsubmit}[1]{}

\title{Consistency Management of Normal Logic Program\\ by Top-down
Abductive Proof Procedure}
\author{Ken Satoh\\Hokkaido University, N13W8, Sapporo, Japan\\
ksatoh@db-ei.eng.hokudai.ac.jp}
\date{}

\begin{document}

\maketitle
\bibliographystyle{named}

\begin{abstract}
This paper presents a method of computing a revision of a
function-free normal logic program. If an added rule is inconsistent
with a program, that is, if it leads to a situation such that no
stable model exists for a new program, then deletion and addition of
rules are performed to avoid inconsistency.  We specify a revision by
translating a normal logic program into an abductive logic program
with abducibles to represent deletion and addition of rules.  To
compute such deletion and addition, we propose an adaptation of our
top-down abductive proof procedure to compute a relevant abducibles to
an added rule.  We compute a minimally revised program, by choosing a
minimal set of abducibles among all the sets of abducibles computed by
a top-down proof procedure.
\end{abstract}

\section{Introduction}
Knowledge base is always subject to change since an environment around
the knowledge base is not guaranteed to be stable forever and even
some error might be included at the initial stage. Therefore, study of
revision of knowledge base is very
important\cite{FUV:Update,Katsuno91,Gardenfors95,Kakas:GSM,Alferes:BR,Witteveen:RVNM,Inoue:AFforTC}.
\cite{FUV:Update} and \cite{Katsuno91} consider a revision of
monotonic theories and there are a lot of researches in this direction
(see~\cite{Gardenfors95} for a survey). \cite{Kakas:UPD} and
\cite{Inoue:AFforTC} consider an update of nonmonotonic
theories to derive a given goal or a given observation. 
\cite{Alferes:BR} and \cite{Witteveen:RVNM} consider a revision
of nonmonotonic theories which is more related to a revision of
monotonic theories studies~\cite{FUV:Update,Katsuno91}; they consider
a revision when inconsistency arises at addition of
rules~\footnote{\cite{Inoue:AFforTC} relate the
latter approach with the former approach by introducing
``anti-explanation of contradiction.''}  In this paper, we follow the
latter approach.

Revision of nonmonotonic theories is especially important for AI,
since it is very rare that commonsense reasoning can be represented as
a monotonic theory.  However, revision of nonmonotonic theories is
more complicated than revision of monotonic theory. In monotonic
theory, if some addition of knowledge or observation leads to
inconsistency, then we can avoid inconsistency by deleting a part of
knowledge base. On the other hand, we might add a piece of assumptions
since deletion leads to inconsistency.

Consider the following program.

$runs(X)\leftarrow car(X),\sim\!\! broken(X).$

$car(c_1)\leftarrow.$

$car(c_2)\leftarrow.$

\vskip 6pt
$\sim$ means ``negation as failure''. The first rule says that if $X$
is a car and $X$ is not known to be broken, $X$ should run. Since
there is no information about $broken(c_1)$ and $broken(c_2)$ in the
current program, $runs(c_1)$ and $runs(c_2)$ are derived.  Suppose,
however, that we add a rule ``$\bot\leftarrow runs(c_1)$'' meaning that car
$a$ does not run~\footnote{$\bot$ means contradiction.}. Then, we have
no stable model, that is, we are in an inconsistent situation. To fix
this inconsistency, we have (at least) two possible ways.
\begin{enumerate}
\item We simply discard the default rule of car $a$:
$$runs(c_1)\leftarrow car(c_1),\sim\!\! broken(c_1).$$
\item We derive $broken(c_1)$ since if we assumed $\sim\!\! broken(c_1)$, then
contradiction would occur and thus, we have a reason to assume $broken(c_1)$.
\end{enumerate}
The first revision is contraction widely used in belief revision of
monotonic theories~\cite{FUV:Update} but the second is special for
nonmonotonic theories such as normal logic programs. In monotonic
theories, addition of formula can not help to restore consistency, but
in nonmonotonic theories addition can help.  This phenomena were
firstly observed in Doyle's justification TMS and he introduced {\em
dependency-directed backtracking}.  Moreover, in monotonic theories,
contraction or deletion of formula can not produce any inconsistency,
whereas in nonmonotonic theories, deletion can cause inconsistency.

Therefore, we need more functions for revision in nonmonotonic
theories than monotonic ones. In this paper, we propose a top-down
procedure using abduction to compute revision for a normal logic
program when there exists no stable model.

Our idea of using abduction for revision is as follows. We introduce
two kinds of abducibles one of which represents a deletion of each
retractable rule and the other of which represents an addition of each
addable rule.  For a retractable rule, we add negation of a
corresponding abducible in the body of the rule so that if an instance
of abducible is assumed then an instance of rule corresponding with
the instance of abducible is no longer applicable. For an addable
rule, we add a corresponding abducible in the body of the rule so that
if an instance of abducible is assumed then an instance of rule
corresponding with the instance of abducible becomes applicable. 

Then, in order to compute such abducibles to specify revision, we show
that we can use a modification of Satoh and Iwayama's query evaluation
procedure on stable models~\cite{Satoh:QEALP} which is a combination
of integrity constraint checking~\cite{Sadri:DBIC} and abductive
procedure~\cite{Kakas:GSM}. In stead of starting with a subprocedure
which show a derivation of positive literals, we start with a
subprocedure for rule consistency checking to derive abducibles to
specify revision. This procedure traverses rules of a program which is
related with addition or deletion of a rule and we guarantee that a
minimal revision can be found by selecting a minimal set of abducibles
among all the sets of abducibles computed by the rule consistency
checking procedure.

\newcommand{\Tpst}{T_{pst}}
\newcommand{\Ttmp}{T_{tmp}}
\newcommand{\Tbck}{T_{bck}}
\newcommand{\Tdel}{T_{del}}
\newcommand{\Tnew}{T_{new}}
\newcommand{\Rnew}{R_{new}}
\newcommand{\vect}[1]{{\bf #1}}

\section{Revision of Normal Logic Program}
Firstly, we define a revision framework as follows. In this paper, we
consider a function-free normal logic program. We use domain closure
axiom and unique name axiom so that constants in the language for a
program are finite and denote distinct objects. We can easily extend
our results to function-free extended logic programs by translating an
extended logic program into a normal logic program proposed
by~\cite{Gelfond91}.

\begin{define}
A {\em rule} $R$ is of the form:
$$H\leftarrow P_1,...,P_j, \sim\! N_1, ..., \sim\! N_h$$
where $H$, $P_1,...,P_j,N_1,...,N_h$ are atoms.

We call $H$ the {\em head} of the rule $R$ denoted as $head(R)$ and
$P_1,...,P_j, \sim\! N_1, ..., \sim\! N_h$ the {\em body} of the rule
denoted as $body(R)$. If $H=\bot$, we sometimes call the rule an
{\em integrity constraint}.

Let $T$ and $\Tbck$ be sets of rules.
A {\em revision framework} ${\cal R}$ is a pair, $\langle T,\Tbck\rangle$
where $T$ can be divided into two sets of rules $\Tpst$ and $\Ttmp$.
We call $\Tpst$ a {\em persistent part} of ${\cal R}$ and $\Ttmp$ a {\em
temporal part} of ${\cal R}$ and $\Tbck$ a {\em backup part} of ${\cal R}$.
\end{define}

$T$ expresses the current logic program which consists of $\Tpst$ and
$\Ttmp$. $\Tpst$ is an unchanged part which should always be satisfied
such as integrity constraints whereas any part of $\Ttmp$ can be
retracted and any part of $\Tbck$ can be added to restore consistency.
Usage of $\Tbck$ is inspired by back-up semantics proposed
by~\cite{Witteveen:RVNM}.

We use stable model semantics for the above program.
\begin{define}
Let $P$ be sets of rules.  We denote a set of ground rules obtained by
replacing all the variables in every rule of $P$ by every element in
the language as $\Pi_{P}$.
\end{define}

\begin{define}
Let $M$ be a set of ground atoms and $\Pi_{P}^{M}$ be the following
program.
\vskip 4pt
$\Pi_{P}^{M}=$
$\{H\leftarrow B_1,...,B_l|
$\par\qquad\qquad$
``H\leftarrow B_1,...,B_l,\sim\!\! A_1,...,\sim\!\! A_h.''\in \Pi_{P}$
% \qquad\qquad
and $A_i\not\in M$ for each $i=1,...,h.\}$
\vskip 4pt\noindent
Let $min(\Pi_{P}^{M})$ be the least model of $\Pi_{P}^{M}$.
A {\em stable model} for a logic program $P$ is $M$ iff
$M=min(\Pi_{P}^{M})$ and $\bot\not\in M$.

We say that $P$ is consistent if $P$ has a stable model.
\end{define}

Now, we define a revised program in a revision framework.

\begin{define}
\label{revisedprogram}
Let ${\cal R}$ be a revision framework, $\langle (\Tpst\cup\Ttmp),\Tbck\rangle$.
Let $\Rnew$ be a rule.
Then, a {\em revised program} w.r.t. ${\cal R}$ and $\Rnew$ is
$(\Tpst\cup\{\Rnew\})\cup(\Pi_{\Ttmp}-O)\cup I$ such that
\begin{itemize}
\item
$O\subseteq \Pi_{\Ttmp}$
\item
$I\subseteq \Pi_{\Tbck}$
\item
$(\Tpst\cup\{\Rnew\})\cup(\Pi_{\Ttmp}-O)\cup I$ is consistent.
\end{itemize}
We say for such $O$ and $I$ that a pair {\em $\langle O,I\rangle$
accomplishes revision of $\Rnew$ to ${\cal R}$}.

A {\em minimally revised program} w.r.t. ${\cal R}$ and $\Rnew$ is 
$(\Tpst\cup\{\Rnew\})\cup(\Pi_{\Ttmp}-O)\cup I$ such that
there is no revised program
$(\Tpst\cup\{\Rnew\})\cup(\Pi_{\Ttmp}-O')\cup I'$ such that
$I'\subset I$ and $O'\subset O$ where $\subset$ is a strict inclusion.
\end{define}

\begin{example}
\label{revision}
Let\\
$\Tpst$ be $\{c(c_1)\leftarrow.\;\; c(c_2)\leftarrow. \}$ and\\
$\Ttmp$ be $\{r(X)\leftarrow c(X),\sim\!\! b(X).\}$ and\\
$\Tbck$ be $\{b(X)\leftarrow c(X),\sim\!\! r(X).\}$ and\\
$\Rnew$ be ``$\bot\leftarrow r(c_1)$''.\\ Then,
$\Pi_{\Ttmp,(\Tpst\cup\{\Rnew\})\cup\Ttmp\cup\Tbck}=$
% \par
$\{r(c_1)\leftarrow c(c_1),\sim\!\! b(c_1).\;\; r(c_2)\leftarrow c(c_2),\sim\!\! b(c_2)\}$,\\
and $\Pi_{\Tbck,(\Tpst\cup\{\Rnew\})\cup\Ttmp\cup\Tbck}=$
% \par
$\{b(c_1)\leftarrow c(c_1),\sim\!\! r(c_1).\;\; b(c_2)\leftarrow c(c_2),\sim\!\! r(c_2)\}$.

For the above revision, we have the following two minimally revised
programs:
\begin{enumerate}
\item
a program accomplished by $(O_1,I_1)$
where $O_1=\{r(c_1)\leftarrow c(c_1),\sim\!\! b(c_1).\}$
and $I_1=\{\}$:
\par\quad
$(\Tpst\cup\{\Rnew\})$ and $r(c_2)\leftarrow c(c_2),\sim\!\! b(c_2).$
\item
a program accomplished by $(O_2,I_2)$
where $O_2=\{\}$
and $I_2=\{b(c_1)\leftarrow c(c_1),\sim\!\! r(c_1).\}:$
\par\quad
$(\Tpst\cup\{\Rnew\})$ and $\Ttmp$ and $b(c_1)\leftarrow c(c_1),\sim\!\! r(c_1).$
\end{enumerate}
There are other non-minimally revised programs, for example, one
accomplished by $\langle O_1,I_2\rangle$ or by $\langle
O_1\cup\{r(c_2)\leftarrow c(c_2),\sim\!\! b(c_2).\},I_2\cup \{b(c_2)\leftarrow
c(c_2),\sim\!\! r(c_2).\}\rangle$.
\end{example}

In the above example, we follow Giordano's approach~\cite{Giordano:TMS}
where contrapositives of default rules are in the back-up part, but
we can actually assume any rules which we think are appropriate
for back-up rules when inconsistency occurs.

To compute a revised program, we use a translation from a
specification to an abductive logic program and compute a consistent
generalized stable model for the translated program which denotes
deletion and addition of rules.

\begin{define}~\cite{Kakas:GSM}.
An {\em abductive framework} is a pair $\langle P,A\rangle$ where
$A$ is a set of predicate symbols, called {\em abducible predicates}
and $P$ is a set of rules each of whose head is not in $A$.
We call a ground atom for a predicate in $A$ {\em an abducible}.
\end{define}

The semantics of abductive framework is based on a generalized stable model
~\cite{Kakas:GSM}.

The following is a definition of a generalized stable model which can
manipulate abducibles in abductive logic programming.

\begin{define}
Let $\langle P,A\rangle$ be an abductive framework and $\Theta$ be a
set of abducibles. A {\em generalized stable model} $M(\Theta)$ is a
stable model of $P\cup \{H\leftarrow|H\in\Theta\}$.

We say that a model $M(\Theta)$ is a {\em generalized stable model
with a minimal set of abducibles} $\Theta$ if there is no generalized stable model
$M(\Theta')$ such that $\Theta'$ is a proper subset of $\Theta$.
\end{define}

Now, we define a translation of a revision framework into an abductive
framework as follows.

\begin{define}
Let ${\cal R}$ be a revision framework $\langle (\Tpst\cup\Ttmp),\Tbck\rangle$.
We firstly give a name to every rule in $\Ttmp$ and $\Tbck$ such as
$$\phi: H\leftarrow P_1,...,P_j, \sim\! N_1, ..., \sim\! N_h.$$
where $\phi$ is a name for the rule.

A {\em translation for a consistency management of ${\cal R}$} (denoted as $\tau({\cal R})$)
is a set of the following translation from ${\cal R}$ to an abductive framework
$\langle P,A\rangle$ where
\begin{itemize}
\item
$A=\{{\phi^-}^{*}|\phi$ is a name of a rule in $\Ttmp\}$
% \\ \hspace*{10pt}
$\cup\{{\phi^+}^{*}|\phi$ is a name of a rule in $\Tbck\}$
\item
We add every rule in $\Tpst$ into $P$.
\item
We translate every rule in $\Ttmp$ with a name $\phi$
$$\phi: H\leftarrow
P_1,...,P_j, \sim\! N_1, ..., \sim\! N_h$$
into the following rule in $P$:
$$H\leftarrow P_1,...,P_j,\sim\! N_1, ..., \sim\! N_h,\sim {\phi^-}^{*}(\vect{x})$$
where $\vect{x}$ is a tuple of variables in the clause.
\item
We translate every rule in $\Tbck$ with a name $\phi$
$$\phi: H\leftarrow
P_1,...,P_j, \sim\! N_1, ..., \sim\! N_h$$
into the following rule in $P$:
$$H\leftarrow P_1,...,P_j,{\phi^+}^{*}(\vect{x}), \sim\! N_1, ..., \sim\! N_h.$$
\end{itemize}
\end{define}

The following shows that an revised program corresponds with
a generalized stable model.
\begin{theorem}
\label{deltheorem}
Let ${\cal R}$ be a revision framework $\langle (\Tpst\cup\Ttmp),\Tbck\rangle$
and $\Rnew$ be an added
clause. $(\Tpst\cup\{\Rnew\})\cup(\Ttmp-\Tdel\cup\Tnew)$ is a (minimally,
resp.) revised program if and only if there is a generalized stable
model of $\tau(\langle (\Tpst\cup\{\Rnew\})\cup\Ttmp,\Tbck\rangle)$ with a
(minimal, resp.) set of abducibles $\Theta$ s.t.
\begin{itemize}
\item
$\Tdel=\{R|({\phi^-}^{*}(\vect{x})\theta)\in\Theta$
\par\hfill
where $\phi$ is a name of $R\in\Ttmp\}$.
\item
$\Tnew=$\\
$\{R\theta|({\phi^+}^{*}(\vect{x})\theta)\in\Theta$
\par\hfill
where $\phi$ is a name of $R\in\Tbck\}$
$\cup$\\
$\{head(R)\leftarrow body(R),\sim(EQ(\theta_1)),...,\sim(EQ(\theta_n))|$
\par\quad
$({\phi^-}^{*}(\vect{x})\theta_i)\in\Theta$ where $\phi$ is a name of $R\in\Ttmp$
\par\quad
and $EQ(\theta_i)=((x_1=(x_1\theta_i))\land...\land (x_k=(x_k\theta_i)))$
\par\quad
and $\vect{x}=\langle x_1,...,x_k\rangle\}$
\end{itemize}

We say that {\em $\Theta$ (minimally, resp.) realizes revision of
$\Rnew$ to ${\cal R}$.}
\end{theorem}

Note that in the above theorem, we delete whole rules related to
inconsistency and then add modified rules with negation of
conjunctions of disequality in the body of the deleted rules.  The
modified rules are logically equivalent to rules in
Definition~\ref{revisedprogram} since we assume domain closure axiom
and unique name axiom.

\begin{example}
\label{translation}
Consider the revision framework in Example~\ref{revision}.
Let us give names to the rules in $\Ttmp$ and $\Tbck$ as follows:

$\phi_1:r(X)\leftarrow c(X),\sim\!\! b(X).$

$\phi_2:b(X)\leftarrow c(X),\sim\!\! r(X).$\\
\par\noindent
Then $\tau(\langle ((\Tpst\cup\{\Rnew\})\cup\Ttmp),\Tbck\rangle)$ is:
\begin{itemize}
\item
$A=\{{\phi_1^-}^{*},{\phi_2^+}^{*}\}$.
\item
$P$ becomes as follows:\\
$c(c_1)\leftarrow.$\\
$c(c_2)\leftarrow.$\\
$\bot\leftarrow r(c_1)$.\\
$r(X)\leftarrow c(X),\sim\!\! b(X),\sim\!{\phi_1^-}^{*}(X).$\\
$b(X)\leftarrow c(X),{\phi_2^+}^{*}(X),\sim\!\! r(X).$\\
\end{itemize}
Then, we have two generalized models with minimal abducibles:
\begin{enumerate}
\item
$\Theta=\{{\phi_1^-}^{*}(c_1)\}$. Then, a minimally
revised program is:
\par\quad
$(\Tpst\cup\{\Rnew\})$ and $r(X)\leftarrow X\not=a, c(X),\sim\!\! b(X).$
\item
$\Theta=\{{\phi_2^+}^{*}(c_1)\}$. Then, a minimally
revised program is:
\par\quad
$(\Tpst\cup\{\Rnew\})$ and $\Ttmp$ and 
$b(c_1)\leftarrow c(c_1),\sim\!\! r(c_1).$
\end{enumerate}
\end{example}

\newcommand{\inl}{\mbox{$\overline{l}$}}
\newcommand{\inp}{\mbox{$\overline{p}$}}
\newcommand{\negl}{\mbox{$\sim\!\! l$}}
\newcommand{\negp}{\mbox{$\sim\!\! p$}}
\newcommand{\negq}{\mbox{$\sim\!\! q$}}
\newcommand{\negr}{\mbox{$\sim\!\! r$}}

\section{Computing Revision by Abduction}
To compute a revision, it is sufficient to compute a generalized
stable model of $\tau(\langle
(\Tpst\cup\{\Rnew\})\cup\Ttmp,\Tbck\rangle)$.  But, if we concern a
minimal revision, we need to compute all the generalized stable models
and then compare sets of abducibles in these models to choose minimal
sets of abducibles. For this purpose, it is desirable to restrict sets
of abducibles to be compared as small as possible.  This can be done
if we compute only revision related to $\Rnew$.  For example, suppose
that some of temporary rules are not relevant to inconsistency of
addition of $\Rnew$.  If we naively compute all the generalized stable
models, then we have to compare all the combination of in/out of
abducibles for these irrelevant rules.

In order to avoid this kind of redundancy, we modify Satoh and
Iwayama's query evaluation procedure on stable
models~\cite{Satoh:QEALP}. Basically, we change the order of
application of subprocedures so that we can use the procedure for
consistency checking.

We impose rules in a revision framework must be {\em
range-restricted}, that is, any variable in a rule $R$ must occur in
$pos(R)$. However, any rule can be translated into range-restricted
form by inserting a new predicate ``$dom$'' describing Herbrand
universe for every non-range-restricted variable in the rule.

Before showing our procedure to compute revision, we need the following
definitions.  Let $l$ be a literal. Then, $\inl$ denotes the
complement of $l$.
\begin{define}
Let $P$ be a logic program.
A set of resolvents w.r.t. a ground literal $l$ and $T$,
$resolve(l,P)$ is the following set of rules:
\vskip 4pt\noindent
$resolve(l,P)=$
\par
$\{(\bot\!\leftarrow\! L_1,...,L_k)\theta|$ $l$ is negative and
\par\qquad\quad
$H\!\leftarrow\! L_1,...,L_k\in P$ and
% \par\qquad\quad
$\inl=H\theta$ by a ground substitution $\theta\}\cup$
\par
$\{(H\!\leftarrow\! L_1,...,L_{i-1},L_{i+1},...,L_{k})\theta|$
\par\qquad\quad
$H\!\leftarrow\! L_1,...,L_k\in P$ and
% \par\qquad\quad
$l=L_i\theta$ by a ground substitution $\theta\}$
\end{define}

\ifsubmit{
\begin{example}
\label{resolvent}
Consider the following program $P$.
\vskip 8pt
$r(X)\leftarrow c(X),\sim\!\! b(X).$\hfill (1)\par
$b(X)\leftarrow c(X),\sim\!\! r(X).$\hfill (2)\par
\vskip 8pt\noindent
Then, $resolve(\negr(c_1),P)$ is a set of the following rules:
\vskip 8pt
$\bot\leftarrow c(c_1),\sim\!\! b(c_1).$\hfill with the head of (1)\par
$b(c_1)\leftarrow c(c_1).$\hfill with the literal in the body of (2)\par
\end{example}
}

\begin{define}
Let $P$ be a logic program.
A set of deleted rules w.r.t. a ground literal $l$ and $P$,
$del(l,P)$, is the following set of rules:
\vskip 4pt\noindent
$del(l,P)= \{(H\!\leftarrow\! L_1,...,L_k)\theta|$
% \par\qquad\quad
$H\!\!\leftarrow\!\! L_1,...,L_k\in P$ and
% \par\qquad\quad
$\inl=L_i\theta$ by a ground substitution $\theta\}$
\end{define}

\ifsubmit{
\begin{example}
Consider the program $P$ in Example~\ref{resolvent}.
Then, $del(b(c_1),P)$ is a set of the following rule:
\vskip 8pt
$r(c_1)\leftarrow c(c_1),\sim\!\! b(c_1).$\hfill (1)\par
\end{example}
}

\begin{define}
Let $P$ be a logic program and $P^{-}$ be an abducible-and-negation-removed program
obtained by removing all integrity constraints in $P$ and
all the negative literals and abducibles in the body of remaining rule
and $min(P^{-})$ be the least minimal model of $P^{-}$.We define a {\em relevant ground program} $\Omega_{P}$ for $P$ as follows:
\vskip 4pt
$\Omega_{P}\!=\!
\{H\!\!\leftarrow\!\!B_1,...,B_k,\sim\!\! A_1,...,\sim\!\! A_m\!\in\!\Pi_{P}|$
% \par\qquad
$B_i\!\in\! min(P^{-})\mbox{ for each }i\!=\!1,...,k.\}$
\end{define}

We briefly explain our procedure.  Our procedure consists of 4
subprocedures, $rule\_con(R,\Delta)$, $derive(p,\Delta)$,
$literal\_con(l,\Delta)$, and $deleted\_con(R,\Delta)$ where $p$ is a
non-abducible atom and $\Delta$ is a set of ground literals already
assumed and $l$ is a ground literal and $R$ is a rule.
$rule\_con(R,\Delta)$, $literal\_con(l,\Delta)$, and
$deleted\_con(R,\Delta)$ return union of $\Delta$ and a set of ground
literals which are assumed during the execution of the subprocedures.
$derive(p,\Delta)$ return the above union and a substitution for $p$
which are made during the execution of $derive(p,\Delta)$.

In the procedure, we have a {\bf select} operation and a {\bf fail}
operation. The {\bf select} operation expresses a nondeterministic
choice among alternatives. The {\bf fail} operation expresses
immediate termination of an execution with failure. Therefore, a
procedure succeeds when its inner calls of subprocedures do not
encounter {\bf fail}.  We say {\em a subprocedure succeeds with (a
substitution $\theta$ and) a set of assumptions $\Delta$} when the
subprocedure successfully returns ($\theta$ and) $\Delta$.

Our procedure firstly starts from
$rule\_con(\Rnew,\{\})$. $rule\_con(\Rnew,\{\})$ checks the
consistency of a rule $\Rnew$ with a program $P\cup\{\Rnew\}$.  We can
show the consistency of addition of $\Rnew$ by showing one of the
following.
\begin{enumerate}
\item
A literal $l$ in $body(\Rnew)$ can be falsified.
To do so, we invoke subprocedure $literal\_con$
for $\inl$.
\item
Every positive literal $p$ in $body(\Rnew)$ can be made true and every
negative and every abducible literal $l$ can be consistently assumed
and $head(\Rnew)$ consistent.  To do so, we invoke subprocedure $derive$
for $p$ and $literal\_con$ for $l$ and $head(\Rnew)$.
\end{enumerate}

The informal specification of the other 3 subprocedures is as follows.
\begin{enumerate}
\item
\label{literalcon}
$literal\_con(l,\Delta)$ checks the consistency of a ground
literal $l$ with $P\cup\{\Rnew\}$ and $\Delta$.
To show the consistency for assuming $l$, we add $l$ to $\Delta$;
then, we check the
consistency of resolvents and deleted rules w.r.t. $l$ and $P\cup\{\Rnew\}$.
\item
$derive(p,\Delta)$ searches a rule $R$ of $p$ in a program $P\cup\{\Rnew\}$ whose
body can be made true with a ground substitution $\theta$ under a set
of assumptions $\Delta$.  To show that every literal in the body can
be made true, we call $derive$ for non-abducible positive literals in the body.
Then, we check the consistency of other literals in the body with $P\cup\{\Rnew\}$
and $\Delta$. Note that because of the range-restrictedness,
other literals in $R$ become ground
after all the calls of $derive$ for non-abducible positive literals.
\item
$deleted\_con(R,\Delta)$ checks if a deletion of $R$ does not cause
any contradictions with $P\cup\{\Rnew\}$ and $\Delta$.
To show the consistency of the
implicit deletion of $R$, it is sufficient to prove that the head of
every ground instance $R\theta$ in
$\Omega_{P\cup\{\Rnew\}}$ can be made either true or false.
\end{enumerate}
\begin{figure}
\baselineskip 14pt
\par\noindent
$rule\_con(R,\Delta)$ $R$: a rule; $\Delta$: a set of literals
\par\noindent
{\bf begin}
\par\quad
$\Delta_0:=\Delta$, $i:=0$
\par\quad
{\bf for} {\bf every} ground rule $R\theta\in \Omega_{P\cup\{\Rnew\cup\{R\}\}}$ {\bf do}
\par\quad
{\bf begin}
\par\qquad
{\bf select} case (a) or case (b)
\par\qquad
(a) {\bf select} $l\in body(R\theta)$
\par\qquad\quad
{\bf if} $l\in pos(R\theta)\cup abd(R\theta)$ and
\par\qquad\qquad
$literal\_con(\inl,\Delta_i)$ succeeds with $\Delta_{i+1}$
\par\qquad\qquad
{\bf then} $i:=i+1$ and {\bf continue}
\par\qquad\quad
{\bf elseif} $l\in neg(R\theta)$ and
\par\qquad\qquad
$derive(\inl,\Delta_i)$ succeeds with $(\varepsilon,\Delta_{i+1})$
\par\qquad\qquad
{\bf then} $i:=i+1$ and {\bf continue}
\par\qquad
(b) $\Delta^{0}_{i}:=\Delta_{i}$, $j:=0$
\par\qquad\quad
{\bf for} {\bf every} $l\in body(R\theta)$ {\bf do}
\par\qquad\quad
{\bf begin}
\par\qquad\qquad
{\bf if} $l\in pos(R\theta)$
\par\qquad\qquad\quad
and $derive(l,\Delta^{j}_i)$ succeeds with ($\varepsilon,\Delta^{j+1}_i$)
\par\qquad\qquad\quad
{\bf then} $j:=j+1$ and {\bf continue}
\par\qquad\qquad
{\bf elseif} $l\in neg(R\theta)\cup abd(R\theta)$
\par\qquad\qquad\quad
and $literal\_con(l,\Delta^{j}_i)$ succeeds with $\Delta^{j+1}_{i}$
\par\qquad\qquad\quad
{\bf then} $j:=j+1$ and {\bf continue}
\par\qquad\quad
{\bf end}
\par\qquad\quad
{\bf if} $literal\_con(head(R\theta),\Delta^{j}_i)$ succeeds
with $\Delta_{i+1}$
\par\qquad\qquad
{\bf then} $i:=i+1$ and {\bf continue}
\par\quad
{\bf end}
\par\quad
{\bf return} $\Delta_{i}$
\par\noindent
{\bf end} ($rule\_con$)

\vskip 8pt\noindent
$literal\_con(l,\Delta)$ $l$:  a ground literal; $\Delta$: a set of literals
\par\noindent
{\bf begin}
\par\quad
{\bf if} $l\in \Delta$ {\bf then} {\bf return} $\Delta$
\par\quad
{\bf elseif} $l=\bot$ or $\inl\in\Delta$ {\bf then} {\bf fail}
\par\quad
{\bf else}
\par\quad
{\bf begin}
\par\qquad
$\Delta_0:=\{l\}\cup\Delta$, $i:=0$
\par\qquad
{\bf for} {\bf every} $R\in resolve(l,P\cup\{\Rnew\})$ {\bf do}
\par\qquad\quad
{\bf if} $rule\_con(R,\Delta_i)$ succeeds with $\Delta_{i+1}$
\par\qquad\qquad
{\bf then} $i:=i+1$ and {\bf continue}
\par\qquad
{\bf for} {\bf every} $R\in del(l,P\cup\{\Rnew\})$ {\bf do}
\par\qquad\quad
{\bf if} $deleted\_con(R,\Delta_i)$ succeeds with $\Delta_{i+1}$
\par\qquad\qquad
{\bf then} $i:=i+1$ and {\bf continue}
\par\quad
{\bf end}
\par\quad
{\bf return} $\Delta_i$
\par\noindent
{\bf end} ($literal\_con$)
\caption{\label{ruleconandliteralcon}
The definition of $rule\_con$ and $literal\_con$}
\end{figure}
\begin{figure}
\baselineskip 14pt
\par\noindent
$derive(p,\Delta)$ $p$: a non-abducible atom; $\Delta$: a set of literals
\par\noindent
{\bf begin}
\par\quad
{\bf if} $p$ is ground and $p\in \Delta$ {\bf then}
{\bf return} $(\varepsilon,\Delta)$
\par\quad
{\bf elseif} $p$ is ground and $\negp\in \Delta$ {\bf then} {\bf fail}
\par\quad
{\bf else}
\par\quad
{\bf begin}
\par\qquad
{\bf select} $R\in P\cup\{\Rnew\}$
\par\qquad\quad
s.t. $head(R)$ and $p$ are unifiable with an mgu
$\theta$
\par\qquad
{\bf if} such a rule is not found {\bf then} {\bf fail}
\par\qquad
$\Delta_{0}:=\Delta$, $\theta_{0}:=\theta$, $B_{0}:=pos(R\theta)$, $i:=0$
\par\qquad
{\bf while} $B_{i}\not=\{\}$ {\bf do}
\par\qquad
{\bf begin}
\par\qquad\quad
take a literal $l$ in $B_{i}$
\par\qquad\quad
{\bf if} $derive(l,\Delta_i)$ succeeds with ($\sigma_{i},\Delta_{i+1}$)
\par\qquad\qquad
{\bf then} $\theta_{i+1}:=\theta_{i}\sigma_{i}$,
$B_{i+1}:=(B_{i}-\{l\})\sigma_{i}$,
\par\qquad\qquad\quad
$i:=i+1$ and {\bf continue}
\par\qquad
{\bf end}
\par\qquad
$\delta:=\theta_{i}$
\par\qquad
{\bf for} {\bf every} $l\in neg(R\delta)\cup abd(R\delta)$ {\bf do}
\par\qquad
{\bf begin}
\par\qquad\quad
{\bf if} $literal\_con(l,\Delta_i)$ succeeds with $\Delta_{i+1}$
\par\qquad\qquad
{\bf then} $i:=i+1$ and {\bf continue}
\par\qquad
{\bf end}
\par\qquad
{\bf if} $literal\_con(p\delta,\Delta_i)$ succeeds with $\Delta'$
\par\qquad\quad
{\bf then} {\bf return} $(\delta,\Delta')$
\par\quad
{\bf end}
\par\noindent
{\bf end} ($derive$)

\vskip 8pt\noindent
$deleted\_con(R,\Delta)$ $R$: a rule; $\Delta$: a set of literals
\par\noindent
{\bf begin}
\par\quad
{\bf if} $l\in \Delta$ {\bf then} {\bf return} $\Delta$
\par\quad
$\Delta_0:=\Delta$, $i:=0$
\par\quad
{\bf for} {\bf every} ground rule $R\theta\in \Omega_{P\cup\{\Rnew\}}$ {\bf do}
\par\quad
{\bf begin}
\par\qquad
{\bf select} case (a) or case (b)
\par\qquad
(a) {\bf if} $derive(head(R\theta),\Delta_{i})$
succeeds with $(\varepsilon,\Delta_{i+1})$
\par\qquad\quad
{\bf then} $i:=i+1$ and {\bf continue}
\par\qquad
(b) {\bf if} $literal\_con(\sim\!\! head(R\theta),\Delta_{i})$
succeeds with
\par\qquad\quad
$\Delta_{i+1}$
\par\qquad\quad
{\bf then} $i:=i+1$ and {\bf continue}
\par\quad
{\bf end}
\par\quad
{\bf return} $\Delta_{i}$
\par\noindent
{\bf end} ($deleted\_con$)
\caption{\label{deriveanddeletedcon}
The definition of $derive$ and $deleted\_con$}
\end{figure}
Now, we describe a complete specification of the subprocedures in
Figure~\ref{ruleconandliteralcon} and
Figure~\ref{deriveanddeletedcon}.  In Figures,
we denote 
a set of non-abducible positive literals, non-abducible negative literals,
and abducibles (either negative or positive) in a rule $R$ as 
$pos(R)$, $neg(R)$ and $abd(R)$ respectively,
and we denotes empty substitution as $\varepsilon$,
and $\theta_{i}\sigma_{i}$ expresses a composition of two
substitutions $\theta_{i}$ and $\sigma_{i}$.

The following theorem on correctness for rule checking
can be derived from correctness on query evaluation procedure
of~\cite{Satoh:QEALP}.
\begin{theorem}
\label{rulecheck}
Let $\langle P,A\rangle$ be a consistent abductive framework.  Suppose
$rule\_con(\Rnew,\{\})$ succeeds for $P$ with $\Delta$, then there is a
generalized stable model $M(\Theta)$ for $\langle P\cup\Rnew,A\rangle$
such that $\Theta$ includes positive abducibles in $\Delta$.
\end{theorem}
par\noindent
% {\bf Proof} By modification of correctness of query evaluation
% in~\cite{Satoh:QEALP}. $\Box$

\vskip 8pt
The above theorem only guarantees that $R$ is consistent with $P$ and
the procedure produces {\em some} abducibles included in a generalized
stable model. To compute revision, however, we must have the stronger
result that $\Delta$ includes all the necessary ${\phi^{+}}^*$'s and
${\phi^{-}}^{*}$'s.  Actually, we can guarantee this by the following
theorem.
\begin{theorem}
Let ${\cal R}$ be a revision framework $\langle(\Tpst\cup\Ttmp),\Tbck\rangle$
such that $\Tpst\cup\Ttmp$ is consistent and $\Rnew$ be an added rule.
Suppose $rule\_con(\Rnew,\{\})$ succeeds for
$\tau({\cal R})$ with $\Delta$, then, a set of positive abducibles in $\Delta$
realizes revision of $\Rnew$ to ${\cal R}$.
\end{theorem}
\par\noindent
% {\bf Proof} We can show that $\Delta$ is a generalized stable model
% for rules which are inspected during the execution of procedure.
% Then, from all the uninspected rules, we can construct a stable model
% where we have a consistent way of assuminng abducibles thanks to the
% consistency of $\Tpst\cup\Ttmp$.  Therefore, positive abducibles in
% $\Delta$ realizes revision of $\Rnew$ to ${\cal R}$.  $\Box$

\vskip 8pt
The following theorem means that if we can search exhaustively in
selecting the rules or cases and there is a generalized stable model
whose abducibles minimally realizes a revision for addition of rule
$\Rnew$, then we can find such a set of abducibles by our procedure.
Note that this property is always guaranteed if a program is a finite
propositional program or has finite constant symbols and no function
symbols.
\begin{theorem}
\label{complete}
Let ${\cal R}$ be a revision framework $\langle(\Tpst\cup\Ttmp),\Tbck\rangle$
and $\Rnew$ be an added rule.  Suppose that every selection of rules
or cases terminates for $rule\_con(\Rnew,\{\})$ with either success or
failure for $\tau({\cal R})$. If
$\Theta$ minimally realizes revision of $\Rnew$ to $T$, then there is a
selection of rules and cases such that $rule\_con(\Rnew,\{\})$ succeeds
with $\Delta$ where a set of abducibles in $\Delta$ 
is equivalent to $\Theta$.
\end{theorem}
% \par\noindent
% {\bf Proof} We firstly define a relevant part of the program for an added rule.
% $\Theta$ cannot have an abducible for an irrelevant part since
% otherwise we can remove the abducible. Therefore, there exists a stable model
% related to $\Theta$ and we restrict relevant part in the stable model, we
% can construct a rule selection and a case selection according to justified
% sequence of the stable model. $\Box$

Note that we cannot guarantee that positive abducibles of every
$\Delta$ always corresponds with a minimal revision. This problem is
inherited from Satoh's procedure in that it is not guaranteed for the
procedure to produce a minimal abducibles.  However, using the
procedure, we can restrict sets of abducibles related with
inconsistency and, thus, considered sets of abducibles to choosing
minimal sets are smaller than sets of abducibles from a naive
calculation of all the generalized stable models.

\begin{example}
Consider the revision framework in Example~\ref{revision}.  The following
are two sequences of main calls of subprocedures for $rule\_con(\Rnew,\{\})$
to $\tau(U)$ shown in Example~\ref{translation}. In the following,
$rc$, $lc$, $dr$ and $dc$ corresponds with $rule\_con$,
$literal\_con$, $derive$ and $deleted\_con$ respectively, and indexes
in the front express a nesting structure of the calls.  Note that
existence of ``$c(c_2)\leftarrow$.'' in $\Tpst$ does not influence these
derivations.
\end{example}
\par\noindent
{\bf Sequence 1}
(for $\langle I_1,O_1\rangle$ in Example~\ref{revision})
\par\noindent
$rc((\bot\leftarrow r(c_1)),\{\})$\nl
1 $lc(\sim\!\! r(c_1),\{\})$\nl
1.1 $rc((\bot\leftarrow c(c_1),\sim\!\! b(c_1),\sim\!\! {\phi_1^{-}}^{*}(c_1)),\{\sim\!\! r(c_1)\})$\nl
1.1.1 $dr(c(c_1),\{\sim\!\! r(c_1)\})$\nl
\qquad$\;$ {\bf select} $c(c_1)\leftarrow. $\nl
1.1.1.2 $lc(c(c_1),\{\sim\!\! r(c_1)\})$\nl
1.1.1.2.1
$rc((r(c_1)\leftarrow \sim\!\! b(c_1),\sim\!\! {\phi_1^{-}}^{*}(c_1)),\{c(c_1),\sim\!\! r(c_1)\})$\nl
1.1.1.2.1.1 $lc(\sim\!\! b(c_1),\{c(c_1),\sim\!\! r(c_1)\})$\nl
1.1.1.2.1.1.1 $rc((r(c_1)\leftarrow c(c_1),\sim\!\! {\phi_1^{-}}^{*}(c_1)),
$\nl\hfill$
\{c(c_1),\sim\!\! b(c_1),\sim\!\! r(c_1)\})$\nl
1.1.1.2.1.1.1.1 $dr({\phi_1^{-}}^{*}(c_1),\{c(c_1),\sim\!\! b(c_1),\sim\!\! r(c_1)\})$\nl
1.1.1.2.1.1.1.1.1 $lc({\phi_1^{-}}^{*}(c_1),\{c(c_1),\sim\!\! b(c_1),\sim\!\! r(c_1)\})$\nl
1.1.1.2.1.1.1.1.1.1 $dc((r(c_1)\leftarrow c(c_1),\sim\!\! b(c_1),\sim\!\! {\phi_1^{-}}^{*}(c_1)),
$\nl\hfill$
\{{\phi_1^{-}}^{*}(c_1),c(c_1),\sim\!\! b(c_1),\sim\!\! r(c_1)\})$\nl
1.1.1.2.1.1.2 $rc((\bot\leftarrow c(c_1),{\phi_2^{+}}^{*}(c_1),\sim\!\! r(c_1)),
$\nl\hfill$
\{{\phi_1^{-}}^{*}(c_1),c(c_1),\sim\!\! b(c_1),\sim\!\! r(c_1)\})$\nl
1.1.1.2.1.1.2.1 $lc(\sim\!\! {\phi_2^{+}}^{*}(c_1),
$\nl\hfill$
\{{\phi_1^{-}}^{*}(c_1),c(c_1),\sim\!\! b(c_1),\sim\!\! r(c_1)\})$\nl
1.1.1.2.1.1.2.1.1 $dc((b(c_1)\leftarrow c(c_1),{\phi_2^{+}}^{*}(c_1),\sim\!\! r(c_1)),
$\nl\hfill$
\{{\phi_1^{-}}^{*}(c_1),c(c_1),\sim\!\! {\phi_2^{+}}^{*}(c_1),\sim\!\! b(c_1),\sim\!\! r(c_1)\})$\nl
1.1.1.2.2 $rc((b(c_1)\leftarrow {\phi_2^{+}}^{*}(c_1),\sim\!\! r(c_1)),
$\nl\hfill$
\{{\phi_1^{-}}^{*}(c_1),c(c_1),\sim\!\! {\phi_2^{+}}^{*}(c_1),\sim\!\! b(c_1),\sim\!\! r(c_1)\})$\nl
1.2 $rc((b(c_1)\leftarrow c(c_1),{\phi_2^{+}}^{*}(c_1)),
$\nl\hfill$
\{{\phi_1^{-}}^{*}(c_1),c(c_1),\sim\!\! {\phi_2^{+}}^{*}(c_1),\sim\!\! b(c_1),\sim\!\! r(c_1)\})$\nl
1.3 $dc((\bot\leftarrow r(c_1)),
$\nl\hfill$
\{\underline{{\phi_1^{-}}^{*}(c_1)},c(c_1),\sim\!\! {\phi_2^{+}}^{*}(c_1),\sim\!\! b(c_1),\sim\!\! r(c_1)\})$\nl

\vskip 8pt\noindent
{\bf Sequence 2}
(for $\langle I_2,O_2\rangle$ in Example~\ref{revision})
\par\noindent
$rc((\bot\leftarrow r(c_1)),\{\})$\nl
1 $lc(\sim\!\! r(c_1),\{\})$\nl
1.1 $rc((\bot\leftarrow c(c_1),\sim\!\! b(c_1),\sim\!\! {\phi_1^{-}}^{*}(c_1)),\{\sim\!\! r(c_1)\})$\nl
1.1.1 $dr(c(c_1),\{\sim\!\! r(c_1)\})$\nl
\qquad$\;$ {\bf select} $c(c_1)\leftarrow. $\nl
1.1.1.2 $lc(c(c_1),\{\sim\!\! r(c_1)\})$\nl
1.1.1.2.1 $rc((r(c_1)\leftarrow \sim\!\! b(c_1),\sim\!\! {\phi_1^{-}}^{*}(c_1)),\{c(c_1),\sim\!\! r(c_1)\})$\nl
1.1.1.2.1.1 $dr(b(c_1),\{c(c_1),\sim\!\! r(c_1)\})$\nl
\qquad\qquad$\;$ {\bf select} $b(c_1)\leftarrow c(c_1),{\phi_2^{+}}^{*}(c_1),\sim\!\! r(c_1).$\nl
1.1.1.2.1.1.2 $dr(c(c_1),\{c(c_1),\sim\!\! r(c_1)\})$\nl
1.1.1.2.1.1.3 $lc({\phi_2^{+}}^{*}(c_1),\{c(c_1),\sim\!\! r(c_1)\})$\nl
1.1.1.2.1.1.3.1 $rc((b(c_1)\leftarrow c(c_1),\sim\!\! r(c_1)),
$\nl\hfill$
\{{\phi_2^{+}}^{*}(c_1),c(c_1),\sim\!\! r(c_1)\})$\nl
1.1.1.2.1.1.3.1.1 $lc(b(c_1),\{{\phi_2^{+}}^{*}(c_1),c(c_1),\sim\!\! r(c_1)\})$\nl
1.1.1.2.1.1.3.1.1.1 $dc((r(c_1)\leftarrow c(c_1),\sim\!\! b(c_1),\sim\!\! {\phi_1^{-}}^{*}(c_1)),
$\nl\hfill$
\{b(c_1),{\phi_2^{+}}^{*}(c_1),c(c_1),\sim\!\! r(c_1)\})$\nl
1.1.1.2.1.1.4 $lc(\sim\!\! r(c_1),\{b(c_1),{\phi_2^{+}}^{*}(c_1),c(c_1),\sim\!\! r(c_1)\})$\nl
1.1.1.2.1.1.5 $lc(b(c_1),\{b(c_1),{\phi_2^{+}}^{*}(c_1),c(c_1),\sim\!\! r(c_1)\})$\nl
1.1.1.2.2 $rc((b(c_1)\leftarrow {\phi_2^{+}}^{*}(c_1),\sim\!\! r(c_1)),
$\nl\hfill$
\{b(c_1),{\phi_2^{+}}^{*}(c_1),c(c_1),\sim\!\! r(c_1)\})$\nl
1.2 $rc((b(c_1)\leftarrow c(c_1),
$\nl\hfill$
{\phi_2^{+}}^{*}(c_1)),\{b(c_1),{\phi_2^{+}}^{*}(c_1),c(c_1),\sim\!\! r(c_1)\})$\nl
1.3 $dc((\bot\leftarrow r(c_1)),
\{b(c_1),\underline{{\phi_2^{+}}^{*}(c_1)},c(c_1),\sim\!\! r(c_1)\})$\nl

\section{Related Work}
There are works on calculation method of
updates~\cite{Kakas:UPD,Inoue:UPD}.
\cite{Kakas:UPD} propose a top-down procedure to compute view updates
in a database for proving a given goal, but it is not applicable to
updating a normal logic program in general. \cite{Inoue:UPD} give a
bottom-up procedure for an update of an acyclic program to explain
given observations in the perfect model of an acyclic program and
therefore, cannot apply to a normal logic program which has multiple
stable models. Moreover, using bottom-up computation would lead to an
irrelevant derivation to an added rule.

In~\cite{Sakama99}, translation from an update
framework~\cite{Inoue:AFforTC} to an extended logic program is
provided~\footnote {Recently,
\cite{Inoue:Unpulished} independently proposes exactly the same
technique as our translation to show correspondence their {\em
extended abduction} and ordinary abduction.}.  Differences between our
translation and their translation are as follows.
\begin{itemize}
\item They give a translation to compute an update to explain
a goal whereas we consider a revision to avoid inconsistency of
addition of a rule.
\item They introduce a new predicate symbol in stead of
abducibles. This makes their translation rather complex.  If we
translate our translated abductive logic program to a new normal logic
program by a method proposed in~\cite{Satoh91}, the new normal logic
program would be the same as their program.
\item 
They consider addition/deletion of the whole rules to derive a given
observation.  That is, instead of considering deletion/addition of
parts of $\Pi_{\Ttmp}$/$\Pi_{\Tbck}$ in
Definition~\ref{revisedprogram}, they propose deletion/addition of
parts of $\Ttmp$/$\Tbck$ .  At least, however, to handle exception of
integrity constraints in software engineering~\cite{Satoh98}, we
believe that our fine-grained approach is better since we would like
to keep consistent part of integrity constraints for further checking
of other data when some instances cause inconsistency. See the
detailed discussion in~\cite{Satoh98}.
\end{itemize}

There are many procedures to compute stable models, generalized stable
models or abduction. If we use a bottom-up procedure for our
translated abductive logic program to compute all the generalized
stable models naively, then sets of abducibles to be compared would be
larger since abducibles of irrelevant temporary rules and addable
rules with inconsistency will be considered. Therefore, it is better to
compute abducibles related with inconsistency. To our knowledge,
top-down procedure which can be used for this purpose is only Satoh
and Iwayama's procedure since we need a bottom-up consistency checking
of addition/deletion of literals during computing abducibles for
revision. This task is similar to integrity constraint checking
in~\cite{Sadri:DBIC} and Satoh and Iwayama's procedure includes
this task.

\section{Conclusion}
In this paper, we propose an abductive top-down procedure to compute a
minimal revised program which traverses only relevant parts of the
program to the added rule.  It is done by translating a revision
framework of a normal logic program into an abductive logic program.

As a future work, we would like to find an efficient method of
computing a minimal revision directly by combining our top-down
procedure and ATMS-like method of memorizing justifications of
revisions.

\section*{Acknowledgments}

This research is partly supported by Grant-in-Aid for Scientific
Research on Priority Areas, ``Principles for Constructing Evolutionary
Software'', The Ministry of Education, Japan. We also thank
the anonymous referees for valuable comments on this paper.

{}

\end{document}